\documentclass[aps,prl,twocolumn,superscriptaddress,showpacs]{revtex4-2}
\usepackage{graphicx}
\usepackage{array}
\usepackage{amsfonts}
\usepackage{amssymb,amsmath,multirow,rotate}
\usepackage{mathrsfs}
\usepackage{booktabs}
\usepackage{threeparttable}
\usepackage{multirow}
\usepackage{epsfig}
\usepackage{threeparttable}
\usepackage{chngpage}
\usepackage{latexsym}
\usepackage{dcolumn}
\usepackage{graphics,graphicx,bm,fleqn,epic,eepic,float}
\usepackage{verbatim}
\usepackage{subfigure}
\usepackage{color}
\usepackage{xr}
\usepackage{float}
\usepackage{listings} 
\usepackage{placeins} 
\usepackage{tikz}
\usepackage{url}
\usepackage{xcolor}
\usepackage{soul}

\definecolor{red}{rgb}{1,0,0}

\definecolor{blue}{rgb}{0,0,1}

\begin{document}

\title{Emergence of a stochastic resonance in machine learning}

\date{\today}

\author{Zheng-Meng Zhai}
\affiliation{School of Electrical, Computer and Energy Engineering, Arizona State University, Tempe, AZ 85287, USA}

\author{Ling-Wei Kong}
\affiliation{School of Electrical, Computer and Energy Engineering, Arizona State University, Tempe, AZ 85287, USA}

\author{Ying-Cheng Lai} \email{Ying-Cheng.Lai@asu.edu}
\affiliation{School of Electrical, Computer and Energy Engineering, Arizona State University, Tempe, AZ 85287, USA}
\affiliation{Department of Physics, Arizona State University,
Tempe, Arizona 85287, USA}

\begin{abstract}

Can noise be beneficial to machine-learning prediction of chaotic systems? Utilizing reservoir computers as a paradigm, we find that injecting noise to the training data can induce a stochastic resonance with significant benefits to both short-term prediction of the state variables and long-term prediction of the attractor of the system. A key to inducing the stochastic resonance is to include the amplitude of the noise in the set of hyperparameters for optimization. By so doing, the prediction accuracy, stability and horizon can be dramatically improved. The stochastic resonance phenomenon is demonstrated using two prototypical high-dimensional chaotic systems. 


\end{abstract}

\maketitle

The interplay between noise and nonlinear dynamics often leads to surprising
phenomena with potentially significant applications and thus has always been 
an active area of interdisciplinary research. It has been well documented that 
noise can be beneficial to applications of dynamical systems, e.g., enhancing 
the response of a nonlinear system to weak periodic signals, through mechanisms
such as stochastic and coherence 
resonances~\cite{BSV:1981,BPSV:1983,MW:1989,MPO:1994,GNCM:1997,GHJM:1998}.
A parallel development in nonlinear dynamics is the important but challenging 
problem of model-free and data-based prediction of chaotic systems~\cite{FS:1987,Casdagli:1989,SGMCPW:1990,KR:1990,GS:1990,Gouesbet:1991,TE:1992,BBBB:1992,CM:1987,Sauer:1994,Sugihara:1994,Parlitz:1996,SSCBS:1996,HKS:1999,Bollt:2000,HKMS:2000,Sello:2001,MNSSH:2001,Smith:2002,Judd:2003,Sauer:2004,TZJ:2007,WYLKG:2011,WLG:2016}.
In general, there are two kinds of forecasting problems: short term and long 
term. In short-term forecasting, the goal is to predict the detailed dynamical 
evolution of the state variables from specific initial conditions, typically 
for a few oscillation cycles (or Lyapunov times). In long term prediction, the 
aim is to generate the attractor of the system with the correct statistical 
behaviors. According to conventional wisdom, for solving the prediction 
problems, typically noise is detrimental. For example, in short-term 
prediction, because of the sensitive dependence on initial conditions, noise 
will make the predicted state evolution diverge exponentially from the true 
one. In long-term prediction, noise can induce the trajectory to cross the 
basin boundary, leading to a wrong attractor.

In this paper, we report the counterintuitive phenomenon that, in model-free
prediction of chaotic systems with machine learning, a certain amount of 
noise can significantly enhance the prediction accuracy and robustness. 
Similar to a stochastic resonance, too little or too much noise is not useful
and may even downgrade the predictive power of the neural machine, but 
an optimal amount of noise can be beneficial. A central issue is then to 
determine the optimal noise level, which we solve using a generalized scheme 
of hyperparameter optimization. To be concrete, we focus on reservoir 
computing~\cite{Jaeger:2001,MNM:2002,JH:2004,MJ:2013} that has become a 
paradigm in machine-learning based prediction of nonlinear dynamical 
systems~\cite{HSRFG:2015,LBMUCJ:2017,PLHGO:2017,DBN:book,PHGLO:2018,Carroll:2018,ZP:2018,GPG:2019,JL:2019b,FJZWL:2020,KKGGM:2020,VPHSGOK:2020,KFGL:2021a,KLNPB:2021,FKLW:2021,KFGL:2021b,Bollt:2021,GBGB:2021} 
and inject noise into the input signal. A reservoir computer contains a number
of hyperparameters and the prediction performance depends strongly on their 
values. Our simulations have revealed that, if the hyperparameters are not 
optimized, noise in the training data can improve to certain extent the 
prediction performance. However, in order to maximize the predictive power of 
a reservoir computer, it is necessary to find the optimal values of the 
hyperparameters, a task that can be accomplished through, e.g., Bayesian 
optimization~\cite{snoek2012practical,YS:2020}.
The key to the emergence of a stochastic resonance is to treat the noise 
amplitude as one of the hyperparameters, i.e., to regard it as an intrinsic 
parameter of the reservoir computer. Bayesian optimization can then yield the 
optimal noise level. We demonstrate using two prototypical high-dimensional 
chaotic systems that noise with the so-determined amplitude can generate more 
accurate, robust and stable predictions in both short and long terms through 
a stochastic resonance. 


The basic principle of reservoir computing and the optimization method are 
described in Supplementary Information (SI)~\cite{SI}. There are six 
hyperparameters to be optimized: the spectral radius $\rho$ of the 
reservoir network, the scaling factor $\gamma$ of the input weights, the 
leakage parameter $\alpha$, the regularization coefficient $\beta$, the link 
connection probability $p$ of the random network in the hidden layer, and the 
noise amplitude $\sigma$. To 
determine the optimal hyperparameter values, we use the \emph{surrogateopt} 
function in Matlab~\cite{surrogate}, a Bayesian optimization procedure, and 
employ a surrogate approximation function to estimate the objective function 
and to find the global minimum through sampling and updating. Specifically, the 
\emph{surrogateopt} algorithm~\cite{surrogate_alg} first samples several 
random points and evaluates the objective function at these trial points. The 
algorithm then creates a surrogate model of the objective function by 
interpolating a radial basis function through all the random trial points. 
From the surrogate function, the algorithm identifies the potential minima 
and samples the points about these minima to update the function. 

We demonstrate the benefits of noise to both short-term and long-term 
prediction using two prototypical chaotic systems: the Mackey-Glass (MG) system 
described by a nonlinear delay differential equation and the spatiotemporal 
chaotic Kuramoto-Sivashinsky (KS) system. 
We use the Bayesian algorithm to obtain the optimal values of the six 
hyperparameters (including the noise amplitude $\sigma$). We then choose
a number of $\sigma$ values away from the optimal value and test the prediction performance. For each such fixed $\sigma$ value, we optimize the other five
hyperparameters. For a different value of $\sigma$, the set of the other five 
hyperparameters is then different. As we will demonstrate, as the noise
amplitude deviates from the optimal value on either side, there is a gradual 
deterioration of the prediction performance, signifying the emergence of a 
stochastic resonance.

\begin{figure} [ht!]
\centering
\includegraphics[width=\linewidth]{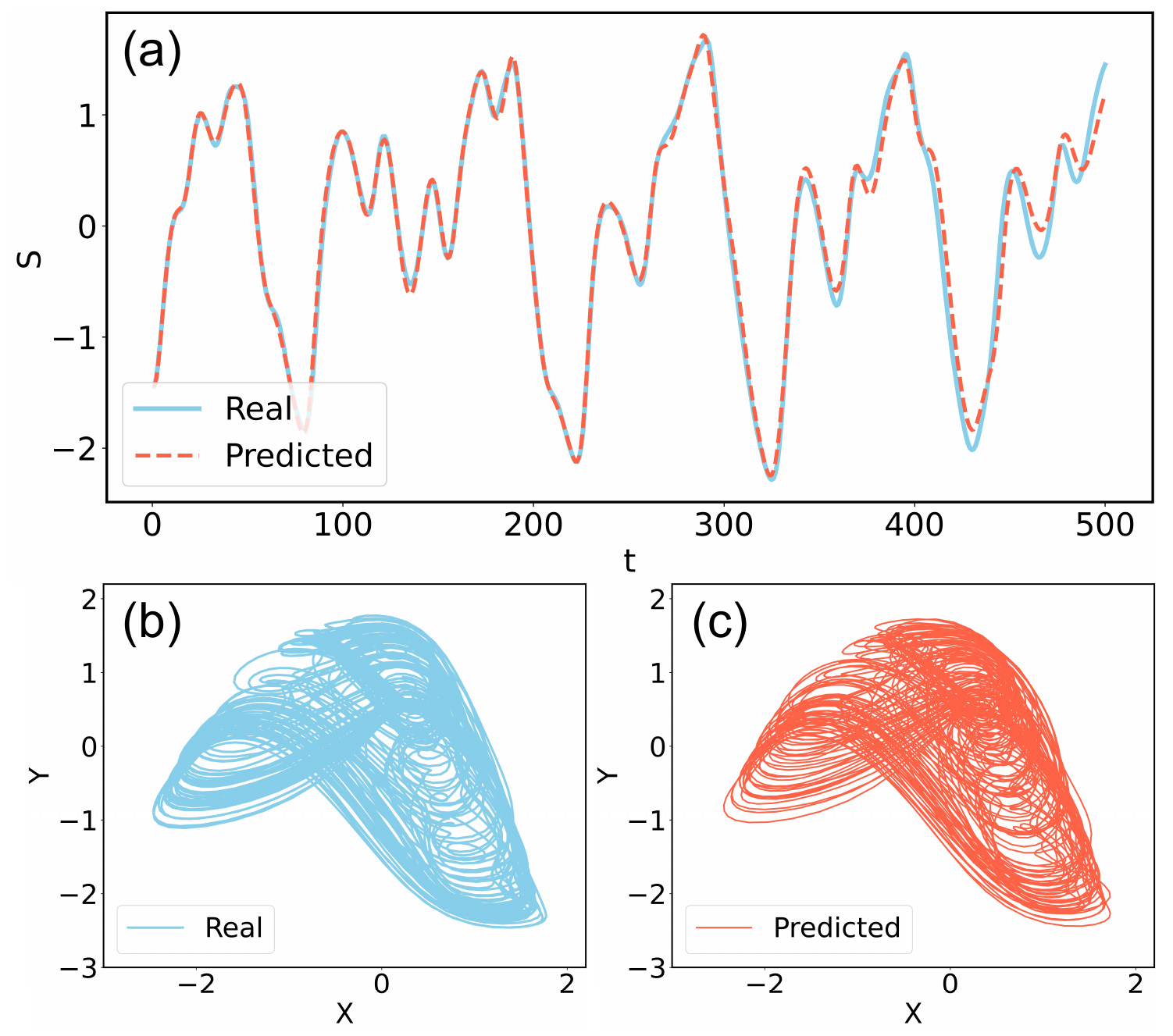} 
\caption{Short-term and long-term prediction of the MG system for 
$\tau=30$. The optimal noise amplitude is $10^{-1.97}$. (a) Machine 
predicted system evolution (red trace) in comparison with the ground truth 
(blue). The predicted state evolution agrees with the true evolution for a 
time period that contains about 15 local maxima ($T = 500 \Delta t$), a result 
that is significantly better than those without optimal noise. 
(b,c) Representation of the true and predicted attractor in the 
$\{X\equiv s(t), Y\equiv s(t-\tau)\}$ plane. The prediction time length is 
$T = 10,000\Delta t$.}
\label{fig:pred_mg}
\end{figure}

Our first example is the MG system~\cite{mackey1977oscillation} described by
$\dot{s}(t)=a s(t-\tau)/\left(1+\left[ s(t-\tau)\right]^c\right)-b s(t)$,
where $\tau$ is the time delay, $a$, $b$ and $c$ are parameters. The state of 
the system at time $t$ is determined by the entire prior state history within 
the time delay, making the phase space of the system infinitely dimensional. 
To be concrete, we use two values of the time delay: $\tau=17$ and $\tau=30$, 
and fix the other three parameters as $a=0.2$, $b=0.1$, and $c=10$. For 
$\tau=17$, the system exhibits a chaotic attractor with one positive Lyapunov 
exponent: $\lambda_+ \approx 0.006$. For $\tau=30$, the system has a chaotic 
attractor with two positive Lyapunov exponents~\cite{wernecke2019chaos}: 
$\lambda_+ \approx 0.011$ and 0.003. 
To generate the one-dimensional MG time series data, we integrate the delay
differential equation with the time step $h=0.01$ and generate the training 
and testing data by sampling the time series every 100 steps: 
$\Delta t = 100h = 1.0$, where $\Delta t$ is evolutionary time step of the 
dynamical network in the hidden layer of the reservoir computer. To remove
any transient behavior, we disregard the first $10,000 \Delta t$ in the 
training dataset. The length of training data is $T = 150,000 \Delta t$. The 
step after the training data marks the start of the testing data, whose length 
depends on whether the task is to make short-term or long-term prediction.
The time series data are pre-processed by using $z$-score normalization: 
$z(t)=\left[s(t)-\bar{s}\right]/\sigma_s$, where $s(t)$ is the original time
series, $\bar{s}$ and $\sigma_s$ are the mean and standard deviation of $s(t)$,
respectively. For $\tau=17$ and $\tau=30$ in the MG system, the testing lengths 
for Bayesian optimization are $T_{\rm opt} = 900 \Delta t$ and $300\Delta t$,
respectively, which are also the testing lengths for short-term prediction. 
The so obtained optimal hyperparameter values are listed in 
Tab.~\ref{tab:opt}. Figure~\ref{fig:pred_mg}(a) shows, for $\tau=30$, 
representative results of short-term prediction of the state evolution, where 
Gaussian noise with the optimal amplitude is injected into the training time 
series. Results of long-term prediction in terms of the attractors in the plane
$\{ X\equiv s(t), Y\equiv s(t-\tau)\}$ are shown in Fig.~\ref{fig:pred_mg}(b). 
Visually and statistically, the predicted attractor cannot be distinguished 
from the true attractor. Prediction results for $\tau=17$ are presented in 
SI~\cite{SI}.

\begin{table}[ht!]
\caption{\label{tab:opt}
Optimal hyperparameter values for MG and KS}
\begin{ruledtabular}
\begin{tabular}{ccccccc}
System & $\rho$ & $\gamma$ & $\alpha$ & $\beta$ & $p$ & $\sigma$\\
\hline 
MG ($\tau=17$) & 1.62 & 0.55 & 0.64 & $10^{-6.0}$ & 0.99 & $10^{-3.42}$\\
MG ($\tau=30$) & 1.27 & 0.23 & 0.57 & $10^{-6.4}$ & 0.09 & $10^{-1.97}$\\
KS & 0.01 & 0.35 & 0.62 & $10^{-9.0}$ & 0.21 & $10^{-2.35}$\\
\end{tabular}
\end{ruledtabular}
\end{table}

Our second example is the one-dimensional KS 
system~\cite{KT:1976,Sivashinsky:1977}, a paradigm not only in physics and 
chemistry but also in applications of reservoir computing for demonstrating 
the predictive power for high-dimensional dynamical systems~\cite{PHGLO:2018}. 
The system equation is 
$\frac{\partial u}{\partial t}+\mu \frac{\partial^4 u}{\partial x^4}+\phi(\frac{\partial^2 u}{\partial x^2}+u\frac{\partial u}{\partial x})=0$,
where $u(x,t)$ is a scalar field defined in the spatial domain $0\leq x\leq L$,
$\mu$ and $\phi$ are parameters. We set $\mu=1$ and $\phi=1$, and use the 
periodic boundary condition. As the domain size $L$ increases, the system 
becomes progressively more high-dimensionally chaotic with the number of 
Lyapunov exponents increasing linearly with the system size~\cite{EBMR:2019}.  
As a representative case of high-dimensional chaos, we choose $L=60$, where 
the system has seven positive Lyapunov exponents: $\lambda_+ \approx 0.089$, 
0.067, 0.055, 0.041, 0.030, 0.005, and 0.003. The length of the training data
is about 1000 Lyapunov times (after disregarding a transient of about 300 
Lyapunov times), where a Lyapunov time is defined as the inverse of the largest
positive exponent. 
The testing data for short-term and long-term prediction are taken immediately
after the training data of six and 100 Lyapunov times, respectively.

\begin{figure}[ht!]
\centering
\includegraphics[width=\linewidth]{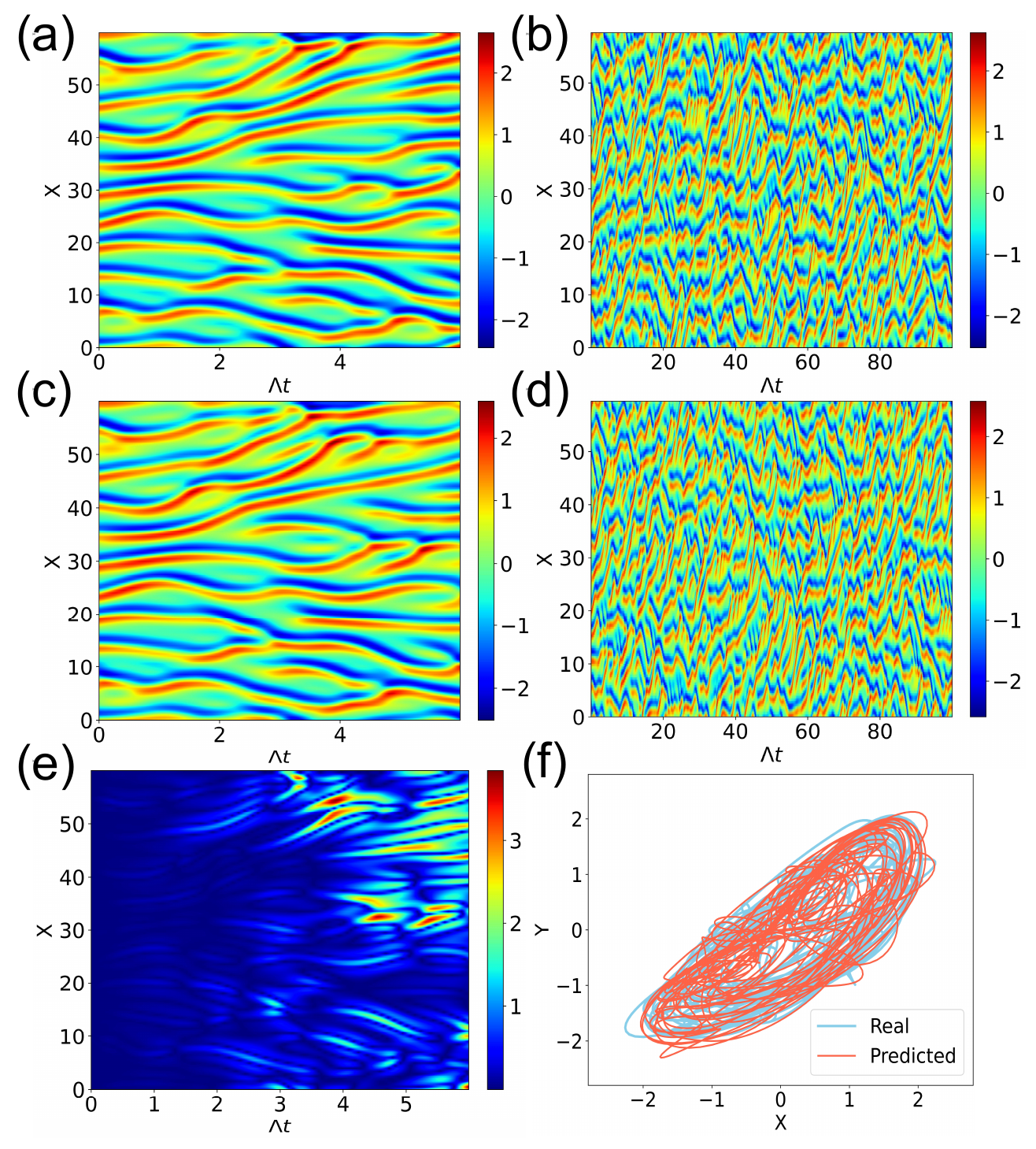} 
\caption{Short-term and long-term prediction of the KS system.
(a,b) True short-term (six Lyapunov times) and long-term (100 Lyapunov 
times) spatiotemporal evolution of the nonlinear field $u(x,t)$, respectively,
(c,d) the predicted field $\hat{u}(x,t)$ in short and long terms, respectively.
(e) Difference between the predicted and true fields defined as 
$D(x,t) \equiv \sqrt{\left[u(x,t)-\hat{u}(x,t)\right]^2}$. (f) Overlapped image
of the true and predicted attractors in terms of the 4th and 5th dimension of 
the KS system. The values of the optimal hyperparameters (including the optimal 
noise amplitude) are listed in Tab.~\ref{tab:opt}.}
\label{fig:pred_ks}
\end{figure}

\begin{figure} [ht!]
\centering
\includegraphics[width=\linewidth]{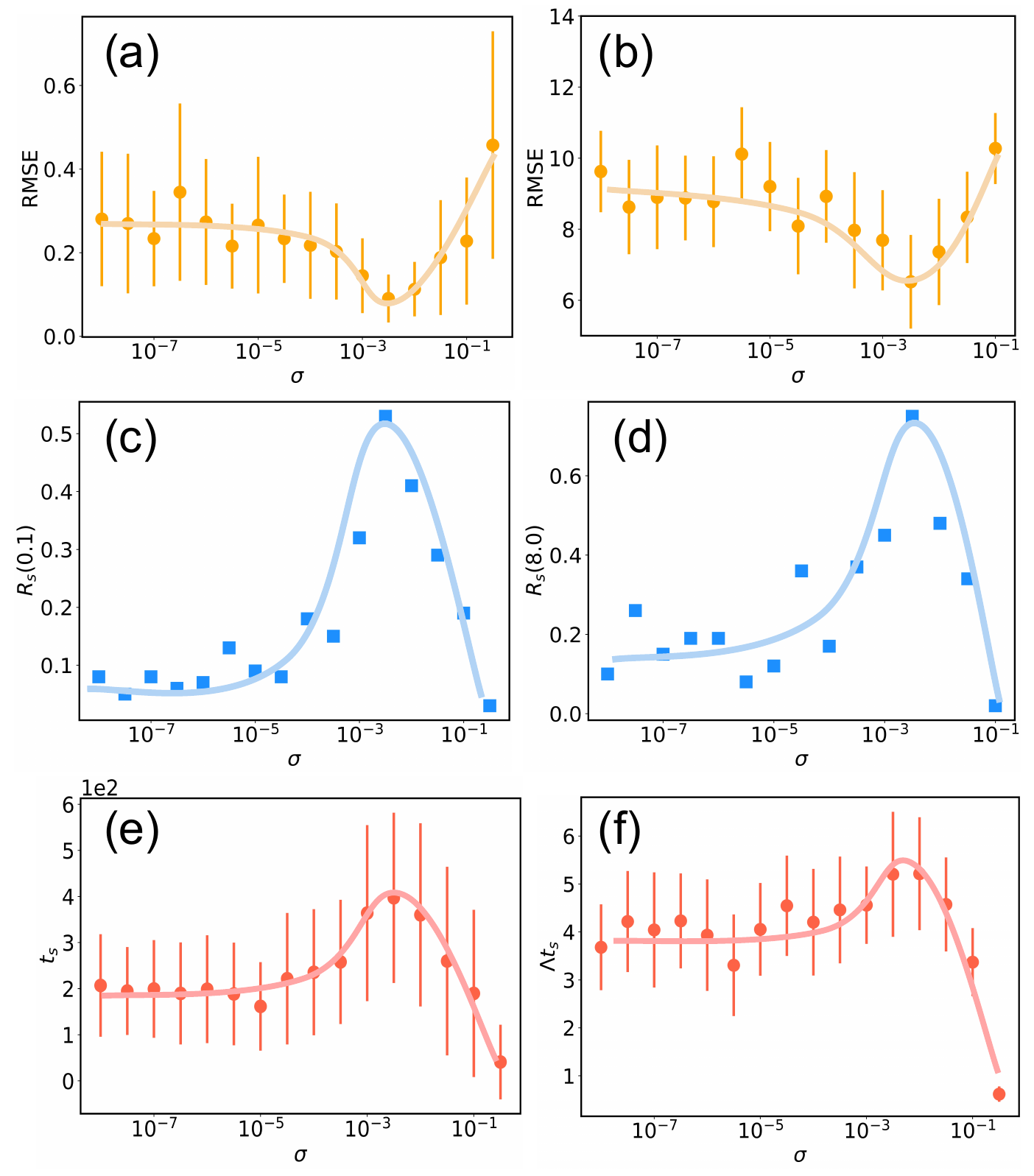} 
\caption{Stochastic resonance associated with short-term prediction of chaotic 
systems. Shown are three measures of short-term prediction versus the noise
amplitude for two examples: left column - MG system for $\tau = 30$ 
($r_c = 0.1$, length of prediction time window $=300 \Delta t$), 
right column - KS 
system ($r_c = 8.0$, length of prediction time window $=$ five Lyapunov times);
top row - RMSE, middle row - prediction stability $R_s(r_c)$, bottom row - 
prediction horizon $t_s$. The error bars are obtained from an ensemble 
of 80 performing reservoir computers. For each chaotic system, a specific and 
unique noise level emerges at which each prediction measure is optimized, 
which is characteristic of a stochastic resonance.}
\label{fig:three_measures}
\end{figure}

\begin{figure} [ht!]
\centering
\includegraphics[width=\linewidth]{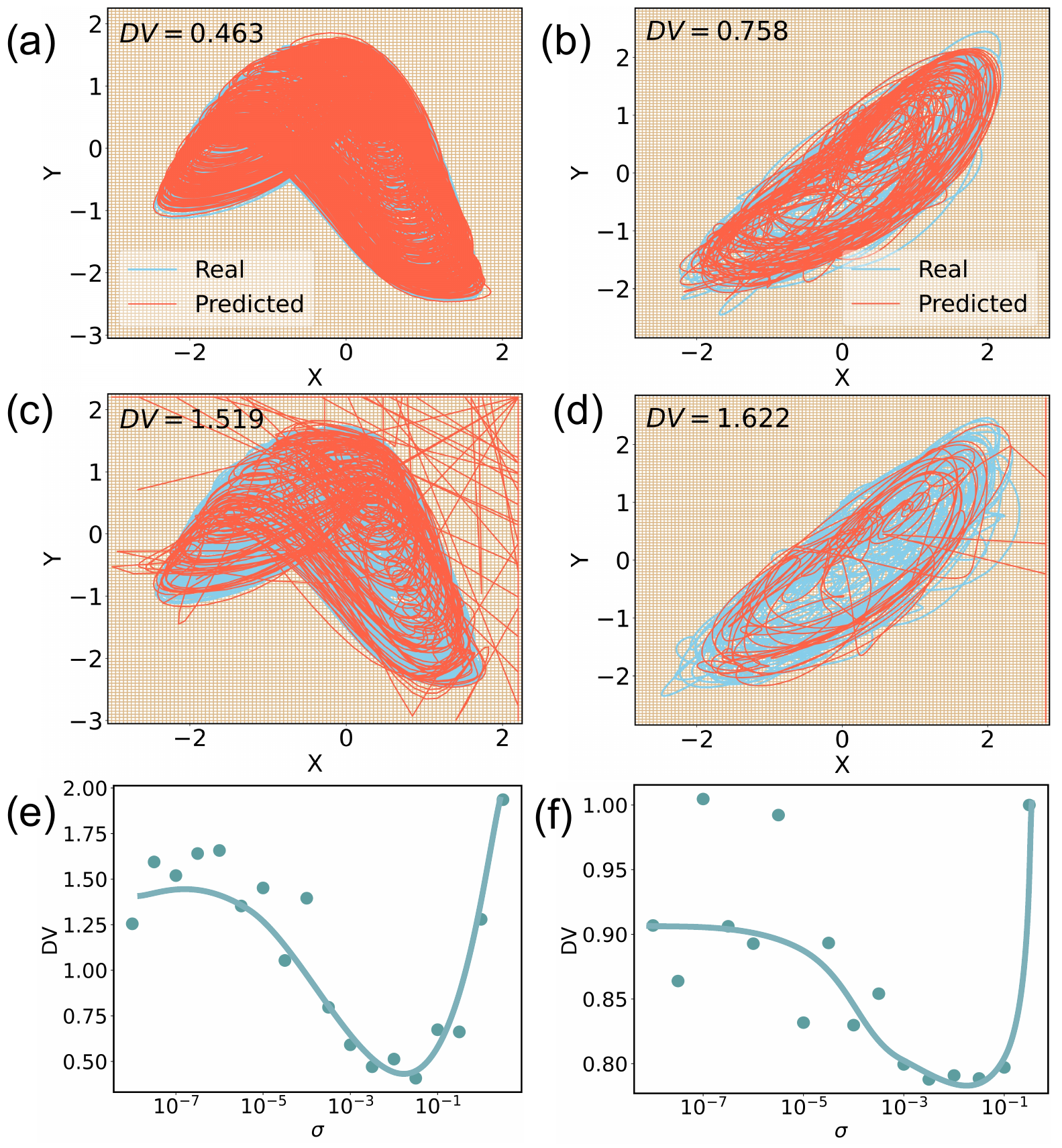} 
\caption{Quantifying long-term prediction through the deviation value DV. 
(a,b) Successful cases of attractor prediction in the presence of optimal 
noise for the MG system for $\tau=30$ and KS system, respectively. 
(c,d) Unsuccessful cases of attractor prediction without noise for the two 
systems. The two-dimensional phase space for the MG system is 
$\{X(t) \equiv s(t), Y(t) \equiv s(t-\tau)\}$. For the KS system, the space 
is $\{X(t) \equiv u(4,t), Y(t) \equiv u(5,t)\}$. (e,f) DV versus the noise 
amplitude for the MG and KS systems, respectively. There exists an optimal 
noise amplitude at which the DV value is minimized, which agrees with the 
optimal noise level determined from the corresponding short-term prediction 
results in Fig.~\ref{fig:three_measures}.}
\label{fig:long_dv}
\end{figure}

Figure~\ref{fig:pred_ks} shows the results of short-term and long-term 
predictions of the KS system. It can be seen that the reservoir computing 
machine with the aid of optimal noise not only can accurately predict the 
short-term spatiotemporal evolution but also is able to replicate the 
long-term attractor with the correct statistical behavior. 

Associated with a conventional stochastic resonance in nonlinear systems, some 
performance measure such as the signal-to-noise ratio is maximized for an 
optimal noise level. Can a similar phenomenon arise in machine-learning 
prediction of chaotic systems, where the prediction performance reaches a 
maximum for an optimal noise level and deteriorates as the noise amplitude 
deviates from the optimal value? That is, how do we ascertain if the optimal 
noise amplitude values from Bayesian optimization as listed in 
Tab.~\ref{tab:opt} are indeed optimal? 

To address these questions, we test the performance when the noise amplitude 
deviates from the optimal value. For any such fixed noise amplitude, the five
other hyperparameters are optimized before training. In particular, we vary
the noise amplitude (uniformly on a logarithmic scale) in the range 
$[10^{-8}, 10^{-0.5}]$. For each noise amplitude, we fix it during the Bayesian
optimization and optimize the other five hyperparameters ($\rho$, $\gamma$, 
$\alpha$, $\beta$, and $k$). For different values of the noise amplitude, the 
so obtained values of the other five hyperparameters are listed in three 
Supplementary tables~\cite{SI}.

We demonstrate the emergence of a stochastic resonance for both short-term and 
long-term predictions. 
To characterize the performance of short-term prediction, besides the 
conventional RMSE (defined in SI~\cite{SI}), we introduce two additional 
measures: prediction horizon and stability, where the former (denoted as $t_s$)
is the maximal time interval during which the RMSE is below a threshold 
and the latter is the probability that a reservoir computer generates stable 
dynamical evolution of the target chaotic system in a fixed time window, which 
is defined as $R_s(r_c)=(1/n) \sum_{i=1}^n \left[{\rm RMSE} < r_c\right]$, 
where $r_c$ is the RMSE threshold, $n$ is the number of iterations, and        
$\left[\cdot\right] = 1$ if the statement inside is true and zero otherwise.


Figure~\ref{fig:three_measures} shows the RMSE, the prediction stability 
$R_s(r_c)$, and the prediction horizon versus the noise amplitude $\sigma$ for
the MG system for $\tau = 30$ (left column, $r_c = 0.1$), as well as the KS
system (right column, $r_c = 8.0$). In both cases, an optimal noise level 
emerges in the sense that a prediction measure versus the noise amplitude 
exhibits either a ``bell shape'' or an ``anti-bell shape'' type of variation 
about an optimal point. Figure~\ref{fig:three_measures} thus provides strong
evidence for a stochastic resonance associated with short-term performance of 
machine-learning prediction of chaotic systems. The results for MG 
for $\tau=17$ are presented in SI~\cite{SI}. 

We now demonstrate the emergence of a stochastic resonance from long-term 
prediction through a quantitative measure that we introduce to characterize
the corresponding performance, as shown in Fig.~\ref{fig:long_dv} for the MG 
system for $\tau=30$ (left column) and the KS system (right column). The 
measure is the deviation value (DV), which characterizes the ability for 
a trained reservoir computer to capture the dynamical ``climate'' of the 
target system (its detailed definition can be found in SI~\cite{SI}). In 
each case, there is an optimal noise amplitude at which the DV value is 
minimized [Figs.~\ref{fig:long_dv}(e) and \ref{fig:long_dv}(f)], which agrees 
with the optimal value of the noise amplitude from the short-term prediction 
results in Fig.~\ref{fig:three_measures}, providing additional support for the 
emergence of a stochastic resonance in machine learning in terms of long-term 
prediction of chaotic attractors. The emergence of such a stochastic resonance 
from long-term prediction for the MG system for $\tau=17$ is treated in 
SI~\cite{SI}. 

To summarize, we have uncovered the emergence of a stochastic resonance in 
machine-learning prediction of chaotic systems. Focusing on reservoir computing,
we find that injecting noise into the training data can be beneficial 
to both short- and long-term predictions. In particular, for short-term 
prediction, a number of characterizing quantities such as the prediction 
accuracy, stability, and horizon can be maximized by an optimal level of noise 
that can be found through hyperparameter optimization. For long-term 
prediction, optimal noise can significantly increase the chance for the machine 
generated trajectory to stay in the vicinity of (or to shadow) the true 
attractor of the target chaotic system. Intuitively, training with noise can 
enhance the machine's tolerance to random fluctuations, which can be beneficial
especially when the target system is chaotic. This argument suggests that the 
optimal noise level should be on the same order of magnitude as the one-step 
prediction error in noiseless prediction, which is indeed so as verified by our
numerical examples. Our work extends the ubiquitous phenomenon of stochastic 
resonance in nonlinear dynamical systems to the realm of machine learning, 
where deliberate noise combined with hyperparameter optimization can be a 
practically feasible approach to enhancing the predictive power of the neural 
machine.

We note that, previously the role of noise in neural network training was 
studied, e.g., adding noise to the training data for convolutional neural 
networks can play the role of regularization to reduce overfitting in the 
learning models~\cite{smilkov2017smoothgrad}. In reinforcement learning, 
injecting noise into the signals can help the system reach the persistent 
excitation condition to facilitate parameter 
estimation~\cite{XYS:2013,KWD:2016}. How noise negatively affects the 
prediction of chaotic systems has recently been considered~\cite{SDG:2021}, 
where long short-term memory machines tend to be more resistant to noise 
than other machine-learning methods. The beneficial role of noise in 
machine-learning prediction has also been recognized~\cite{bishop1995training,grandvalet1997noise,wyffels2008stable,jim1996analysis}. 
In spite of the previous efforts, to our knowledge, the interplay between 
noise and machine-learning prediction of dynamical systems was not 
systematically studied prior to our work. The discovery of stochastic 
resonance in machine learning fills this gap.

Finally, we remark that, in the vast literature on stochastic resonance, the 
paradigmatic model of mechanical motion of a particle in a double-well 
potential subject to stochastic forcing is often used to explain the observed 
resonance phenomenon. However, it is difficult to apply this model to our 
machine-learning system, as the dynamics of the high-dimensional neural 
network in the hidden layer of a reservoir computer are extraordinarily 
complicated. Nonetheless, the predictive ability of the reservoir computer 
can be related to generalized synchronization (see Sec.~IV in SI~\cite{SI}). 
Previous works in the past two decades demonstrated that noise can induce and 
enhance synchronization in nonlinear and complex dynamical 
systems~\cite{TMHP:2001,HSS:2003,GLLG:2006,WLZ:2007,NAK:2008,ESS:2012,LP:2013,KIHN:2014,RS:2018,TPVE:2020,SADKS:2021}.
We speculate that the mechanism responsible for the stochastic resonance
phenomenon reported here is noise-enhance generalized synchronization.

This work was supported by AFOSR under Grant No.~FA9550-21-1-0438.


%
\end{document}